\documentclass[runningheads,a4paper]{llncs}

\usepackage{amssymb}
\usepackage{graphicx}
\usepackage{mathtools}
\usepackage{url}

\begin{document}

\mainmatter  

\title{Surgical Activity Recognition in Robot-Assisted Radical Prostatectomy using Deep Learning}

\titlerunning{RARP Activity Recognition}

%
%
\author{Aneeq Zia\textsuperscript{1}, Andrew Hung\textsuperscript{2}, Irfan Essa\textsuperscript{1}, and Anthony Jarc\textsuperscript{3}}
\authorrunning{Aneeq Zia, Andrew Hung, Irfan Essa, and Anthony Jarc}

\institute{\textsuperscript{1} Georgia Institute of Technology, Atlanta, GA, USA\\
\textsuperscript{2} University of Southern California, Los Angeles, CA, USA \\
\textsuperscript{3} Medical Research, Intuitive Surgical Inc., Norcross, GA, USA}

\toctitle{Lecture Notes in Computer Science}
\tocauthor{Authors' Instructions}
\maketitle

\begin{abstract}
Adverse surgical outcomes are costly to patients and hospitals. Approaches to benchmark surgical care are often limited to gross measures across the entire procedure despite the performance of particular tasks being largely responsible for undesirable outcomes. 
In order to produce metrics from tasks as opposed to the whole procedure, methods to recognize automatically individual surgical tasks are needed. In this paper, we propose several approaches to recognize surgical activities in robot-assisted minimally invasive surgery using deep learning. We collected a clinical dataset of 100 robot-assisted radical prostatectomies (RARP) with 12 tasks each and propose \textit{`RP-Net'}, a modified version of InceptionV3 model, for image based surgical activity recognition. We achieve an average precision of 80.9\% and average recall of 76.7\% across all tasks using RP-Net which out-performs all other RNN and CNN based models explored in this paper. 
Our results suggest that automatic surgical activity recognition during RARP is feasible and can be the foundation for advanced analytics.
\end{abstract}

\section{Introduction}

Adverse outcomes are costly to the patient, hospital, and surgeon. Although many factors contribute to adverse outcomes, the technical skills of surgeons are one important and addressable factor. Virtual reality simulation has played a crucial role to train and improve the technical skills of surgeons, however, intraoperative assessment has been limited to feedback from attendings and/or proctors. Aside from the qualitative feedback from experienced surgeons, quantitative feedback has remained abstract to the level of an entire procedure, such as total duration. Performance feedback for one particular task within a procedure might be more helpful to direct opportunities of improvement. Similarly, statistics from the entire surgery may not be ideal to show an impact on outcomes. For example, one might want to closely examine the performance of a single task if certain adverse outcomes are related to only that specific step of the entire procedure \cite{hung2018utilizing}. Scalable methods to recognize automatically when particular tasks occur within a procedure are needed to generate these metrics to then provide feedback to surgeons or correlate to outcomes.

The problem of surgical activity recognition has been of interest to many researchers. Several methods have been proposed to develop algorithms that automatically recognize the phase of surgery. For laparoscopic surgeries, \cite{twinanda2017endonet} proposed \textit{`Endo-Net'} for recognizing surgical tools and phases in cholecystectomy using endoscopic images. In \cite{dipietro2016recognizing}, RNN models were used to recognize surgical gestures and maneuvers using kinematics data. In \cite{zia2017temporal}, unsupervised clustering methods were used to segment training activities on a porcine model. In \cite{padoy2012statistical}, hidden markov models were used to segment surgical workflow within laparoscopic cholecystectomy.

In this work, we developed models to detect automatically the individual steps of robot-assisted radical prostatectomies (RARP). Our models break a RARP into its individual steps, which will enable us to provide tailored feedback to residents and fellows completing only a portion of a procedure and to produce task-specific efficiency metrics to correlate to certain outcomes. By examining real-world, clinical RARP data, this work builds foundational technology that can readily translate to have direct clinical impact.\\

\noindent\textit{\textbf{Our contributions are}}, (1) a detailed comparison of various deep learning models using image and robot-assisted surgical system data from clinical robot-assisted radical prostatectomies; (2) RP-Net, a modified InceptionV3 architecture that achieved the highest surgical activity recognition performance out of all models tested; (3) a simple median filter based post processing step for significantly improving procedure segmentation accuracies of different models.

\section{Methodology}

\begin{figure}[t]
	\centering
	\includegraphics[width=1.0\columnwidth]{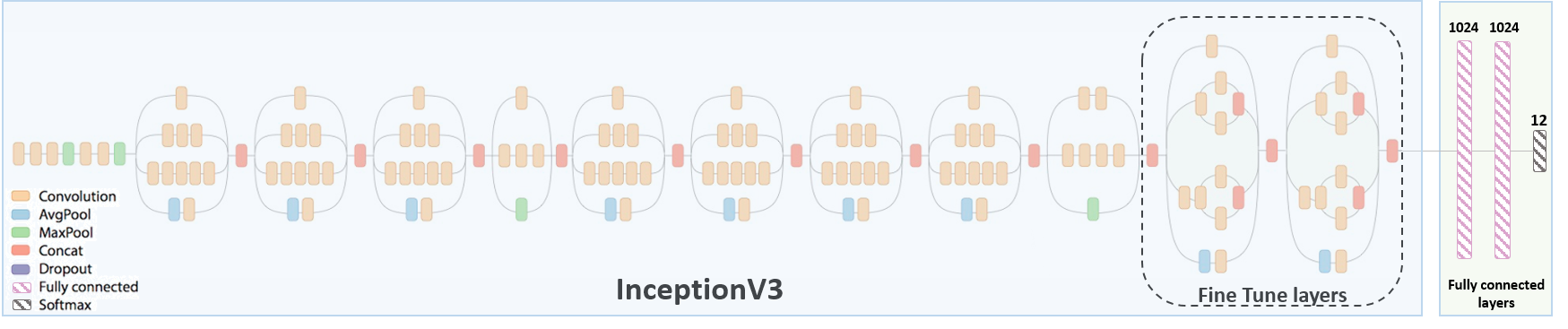}
	\caption{RP-Net architecture. The portion shown in blue is the same as InceptionV3 architecture, whereas the green portion shows the fully connected (fc) layers we add to produce RP-Net. The number of units for each fc layers is also shown. Note the last two layers of InceptionV3 are fine-tuned in RP-Net.}
	\label{fig:model}
\end{figure}

\noindent The rich amount of data that can be collected from the da Vinci (dV) surgical system (Intuitive Surgical, Inc., Sunnyvale, CA USA) enables multiple ways to explore recognition of the type of surgical tasks being performed during a procedure. Our development pipeline involves the following steps: (1) extraction of endoscopic video and dV surgical system data (kinematics and a subset of events), (2) design of deep learning based models for surgical task recognition, and (3) design of post-processing models to filter the initial procedure segmentation output to improve performance. We provide details on modeling below and on our dataset in the next section.\\

\noindent\textit{\textbf{System data based models:}}
The kind of hand and instrument movements surgeons make during procedures can be very indicative of what types of task they are performing. For example, a dissection task might involve static retraction and blunt dissection through in and out trajectories, whereas a suturing task might involve a lot of curved trajectories. Therefore, models that extract motion and event based features from dV surgical system data seem appropriate for task/activity recognition. We explore multiple Recurrent Neural Network (RNN) models using only system data given the recent success of RNNs to incorporate temporal sequences. Since there are multiple data streams coming from the dV surgical system, we employ two types of RNN architectures - \textit{single stream} (SS) and \textit{multi-stream} (MS). For SS, all  data streams are concatenated together before feeding them into a RNN. Whereas, for MS, each data stream is fed into individual RNNs after which the outputs of each RNN are merged together using a fully-connected layer to produce predictions. For training both architecture types, we divide our procedure data into windows of length $W$. At test time, individual windows of the procedure are classified to produce the output segmentation.

\noindent\textit{\textbf{Video based models:}} Apart from the kind of motions a surgeon makes, a lot of task representative information is available in the endoscopic video stream. Tasks which are in the beginning could generally look more \textit{`yellow'} due to the fatty tissues, whereas tasks during the later part of the surgery could look much more \textit{`red'} due to the presence of blood after dissection steps. Moreover, the type and relative location of tools present in the image can also be very indicative of the step that the surgeon is performing. Therefore, we employ various image based convolutional neural networks (CNN) for recognizing surgical activity using video data. Within the CNNs domain, there are two type of CNN architectures that are popular and have been proved to work well for the purpose of recognition. The first type uses single images only with two-dimensional (2D) convolutions in the CNN architectures. Examples of such networks include VGG \cite{Simonyan14c}, ResNet \cite{he2016deep} and InceptionV3 \cite{szegedy2016rethinking}. The second type of architecture uses a volume of images as input (e.g., 16 consecutive frames from the video) and employs three-dimensional (3D) convolutions instead of 2D. C3D is an example of such model \cite{tran2015learning}. A potential advantage of 3D models is that they can learn spatio-temporal features from video data instead of just spatial features. However, this comes at the cost of requiring more data to train as well as longer overall training times. For our task of surgical activity recognition, we employ both types of CNN models and also propose \textit{'RP-Net'} (Radial Prostatectomy Net), which is a modified version of InceptionV3 as shown in Figure \ref{fig:model}.

\noindent\textit{\textbf{Post-Processing:}} Since there are parts of various tasks that are very similar visually and in terms of motions the surgeon is making, the predicted procedure segmentation can have \textit{`spikes'} of mis-classifications. However, it can be assumed that the predicted labels would be consistent within a small window. Therefore, in order to remove such noise from the output, we employ a simple running window median filter of length $F$ as a post-processing step. For corner cases, we append the start and end of the predicted sequence with the median of first and last window of length $F$, respectively, in order to avoid mis-classifications of the corner cases by appending zeros.

\section{Experimental Evaluation}

\noindent\textit{\textbf{Dataset:}} Our dataset consisted of 100 robot-assisted radical prostatectomies (RP) completed at an academic hospital. The majority of procedures were completed by a combination of residents, fellows, and attending surgeons. Each RP was broken into approximately 12 standardized tasks. The order of these 12 tasks varied slightly based on surgeon preference. The steps of each RP were annotated by one resident. A total of 1195 individual tasks were used. Table \ref{dataset} shows general statistics of our dataset.

Each RP recording included one channel of endoscopic video, dV surgical system kinematic data (e.g., joint angles, endpoint pose) collected at 50Hz, and dV surgical system event data (e.g., camera movement start/stop, energy application on/off).

The dV surgical system kinematic data originated from the surgeon console (SSC) and the patient side cart (SI). For both the SSC and SI, the joint angles for each manipulandum and the endpoint pose of the hand controller or instrument were used.  In total, there were 80 feature dimensions for SSC and 90 feature dimensions for SI. The dV surgical system event data (EVT) consisted of many events relating to surgeon interactions with the dV surgical system originating at the SSC or SI. In total, there were 87 feature dimensions for EVT.\\

\begin{table}[t]
	\centering
	\caption{Dataset: the 12 steps of robot-assisted radical prostatectomy and general statistics.}
	\label{dataset}
	\begin{tabular}{|c|c|c|c|}
		\hline
		Task no & Task Name                         & Mean time (sec) & Number of samples \\ \hline
		T1       & mobilize colon / drop bladder     &  1063.2         &       100            \\ \hline
		T2       & Endopelvic fascia           &     764.2      &     98              \\ \hline
		T3       & Anterior bladder neck dissection  &   164.9        &     98              \\ \hline
		T4       & Posterior bladder neck dissection &    617.5       &      100             \\ \hline
		T5       & Seminal vesicles                  &    686.8       &      100             \\ \hline
		T6       & Posterior plane / Denonvilliers   &  171.2         &    99               \\ \hline
		T7       & Predicles / nerve sparing         &    510.6       &     100              \\ \hline
		T8       & Apical dissection                 &    401.1       &      100             \\ \hline
		T9       & Posterior anastomosis             &   403.1        &     100              \\ \hline
		T10      & Anterior anastomosis              &    539.7       &    100               \\ \hline
		T11      & Lymph node dissection Left           &   999.6        &     100              \\ \hline
		T12      & Lymph node dissection Right           &    1103.6       &    100               \\ \hline
	\end{tabular}
\end{table}

\noindent\textit{\textbf{Data preparation:}} Several pre-processing steps were implemented. The endoscopic video was downsampled to 1 frame per second (fps) resulting in 1.4 million images in total. Image resizing and rescaling was model specific. All kinematic data was downsampled by a factor of 10 (from 50Hz to 5Hz). Different window lengths (in terms of the number of samples) $W$ (50, 100, 200 and 300) were tried for training the models and $W=200$ performed the best. We used zero overlap when selecting windows for both training and testing. Mean normalization was applied to all feature dimensions for the kinematic data. All events from the dV surgical system data that occured within each window $W$ were used as input for to our models. The events were represented as a unique integers with corresponding timestamps.\\

\noindent\textit{\textbf{Model training and parameter selection:}} For RNN based models, we implemented both SS and MS architectures for all possible combinations of the three data streams (SSC, SI, and EVT).  Estimation of model hyperparameters was done via a grid search on the number of hidden layers (1 or 2), type of RNN unit (Vanilla, GRU or LSTM), number of hidden units per layer (8, 16, 32, 64, 128, 256, 512 or 1024) and what dropout ratio to use (0, 0.2 or 0.5). For each parameter set, we also compared forward and bi-directional RNN. The best performances were achieved using single layered bi-directional RNNs with 256 LSTM units and a dropout ratio of 0.2. Hence, all RNN based results presented were evaluated using these parameters for SS and MS architecture types.

In CNN based models, we used two approaches - training the networks from randomly initialized weights and fine-tuning the networks from pre-trained weights. For all models, we found that fine-tuning was much faster and achieved better accuracies. For single image based models, we used ImageNet \cite{deng2009imagenet} pretrained weights while for C3D we used Sports-1M \cite{karpathy2014large} pretrained weights. We found that fine-tuning several of the last convolutional layers led to the best performances across models. For the proposed RP-Net, the last two convolutional modules were fine-tuned (as shown in Figure \ref{fig:model}) and the last fully connected layers were trained from random initialization. 

For both RNN- and CNN-based models, the dataset was split to include 70 procedures for training, 10 procedures for validation, and 20 procedures for test.

For the post-processing step, we evaluated performances of all models for values of $F$ (median filter length) ranging from 3 to 2001, and choose a window length that led to maximum increase in model performance across different methods. The final value of  $F$ was set to 301. All parameters were selected based on the validation accuracy.

\noindent\textit{\textbf{Evaluation Metrics:}}  For a given series of ground truth labels $G \in \Re^{N}$ and predictions $P \in \Re^N$, where $N$ is the length of a procedure, we evaluate multiple metrics for comparing the performance of various models.  These include average precision (AP), average recall (AR) and Jaccard index. Precision is evaluated using $P=\frac{tp}{tp+fp}$, recall using $R=\frac{tp}{tp+fn}$ and Jaccard index using $J=\frac{tp}{tp+fp+fn}$, where $tp$, $fp$ and $fn$ represent the true positives, false positives and false negatives, respectively.

\section{Results and Discussion}

The evaluation metrics for all models are shown in Table \ref{results_table}. RP-Net achieved the highest scores across all evaluation metrics out of all models (see last row in Table \ref{results_table}). In general, we observed that the image-based CNN models (except for C3D) performed better than the RNN models. Within LSTM models, MS architecture performed slightly better than SS with the SSC+EVT combination achieving the best performance. For nearly all models, post-processing significantly improved task recognition performance.  

\begin{table}[t]
	\centering
	\caption{Surgical procedure segmentation results using different models. Each cell shows the average metric values across all procedures and tasks in the test set with standard deviations using the original predictions and filtered predictions in the form $ original$ $\vert$ $filtered$. For LSTM models, the modalities used are given in square brackets while the architecture type used is given in parentheses. }
	\label{results_table}
	\resizebox{\textwidth}{!}{
		\begin{tabular}{cccc}
			\hline
			\multicolumn{1}{|c|}{Model Type}                & \multicolumn{1}{c|}{Average Precision}                             & \multicolumn{1}{c|}{Average Recall}                                & \multicolumn{1}{c|}{Average Jaccard Index}                          \\ \hline
			\multicolumn{1}{|c|}{LSTM $\lbrack$ssc+si$\rbrack$ (MS)}                & \multicolumn{1}{c|}{0.585$\pm$0.19 $\vert$ 0.595$\pm$0.21} & \multicolumn{1}{c|}{0.565$\pm$0.21 $\vert$ 0.572$\pm$0.21} & \multicolumn{1}{c|}{0.629$\pm$0.18 $\vert$ 0.645$\pm$0.19}  \\ \hline
			\multicolumn{1}{|c|}{LSTM$\lbrack$ssc+si$\rbrack$ (SS)}                & \multicolumn{1}{c|}{0.559$\pm$0.14 $\vert$ 0.578$\pm$0.15} & \multicolumn{1}{c|}{0.526$\pm$0.16 $\vert$ 0.551$\pm$0.16} & \multicolumn{1}{c|}{0.582$\pm$0.16 $\vert$ 0.606$\pm$0.17}  \\ \hline
			\multicolumn{1}{|c|}{LSTM$\lbrack$ssc+evt$\rbrack$ (MS)}               & \multicolumn{1}{c|}{0.625$\pm$0.13 $\vert$ 0.648$\pm$0.13} & \multicolumn{1}{c|}{0.572$\pm$0.16 $\vert$ 0.593$\pm$0.17} & \multicolumn{1}{c|}{0.633$\pm$0.18 $\vert$ 0.662$\pm$0.19}  \\ \hline
			\multicolumn{1}{|c|}{LSTM$\lbrack$ssc+evt$\rbrack$ (SS)}               & \multicolumn{1}{c|}{0.625$\pm$0.13 $\vert$ 0.641$\pm$0.13} & \multicolumn{1}{c|}{0.567$\pm$0.21 $\vert$ 0.593$\pm$0.22} & \multicolumn{1}{c|}{0.625$\pm$0.18 $\vert$ 0.651$\pm$0.19}  \\ \hline
			\multicolumn{1}{|c|}{LSTM$\lbrack$ssc+si+evt$\rbrack$ (MS)}             & \multicolumn{1}{c|}{0.437$\pm$0.29 $\vert$ 0.458$\pm$0.31} & \multicolumn{1}{c|}{0.226$\pm$0.31 $\vert$ 0.471$\pm$0.32} & \multicolumn{1}{c|}{0.552$\pm$0.15 $\vert$ 0.582$\pm$0.16}  \\ \hline
			\multicolumn{1}{|c|}{LSTM$\lbrack$ssc+si+evt$\rbrack$ (SS)}           & \multicolumn{1}{c|}{0.544$\pm$0.13 $\vert$ 0.579$\pm$0.12} & \multicolumn{1}{c|}{0.518$\pm$0.17 $\vert$ 0.546$\pm$0.17} & \multicolumn{1}{c|}{0.575$\pm$0.15 $\vert$ 0.603$\pm$0.17}  \\ \hline
			\multicolumn{1}{|c|}{InceptionV3}        & \multicolumn{1}{c|}{0.662$\pm$0.12 $\vert$ 0.782$\pm$0.14} & \multicolumn{1}{c|}{0.642$\pm$0.15 $\vert$ 0.759$\pm$0.17} & \multicolumn{1}{c|}{0.666$\pm$0.07 $\vert$ 0.786 $\pm$0.08} \\ \hline
			\multicolumn{1}{|c|}{VGG-19}              & \multicolumn{1}{c|}{0.549$\pm$0.16 $\vert$ 0.695$\pm$0.19} & \multicolumn{1}{c|}{0.481$\pm$0.2 $\vert$ 0.573$\pm$0.22}  & \multicolumn{1}{c|}{0.529$\pm$0.08 $\vert$ 0.634$\pm$0.11}  \\ \hline
			\multicolumn{1}{|c|}{ResNet}           & \multicolumn{1}{c|}{0.621$\pm$0.1 $\vert$ 0.713$\pm$0.12}  & \multicolumn{1}{c|}{0.582$\pm$0.21 $\vert$ 0.673$\pm$0.25} & \multicolumn{1}{c|}{0.622$\pm$0.07 $\vert$ 0.728$\pm$0.08}  \\ \hline
			\multicolumn{1}{|c|}{C3D}          & \multicolumn{1}{c|}{0.442$\pm$0.17 $\vert$ 0.352$\pm$0.21} & \multicolumn{1}{c|}{0.417$\pm$0.19 $\vert$ 0.367$\pm$0.24} & \multicolumn{1}{c|}{0.504$\pm$0.06 $\vert$ 0.418$\pm$0.12}  \\ \hline
			\multicolumn{1}{|c|}{RP-Net} & \multicolumn{1}{c|}{\textbf{0.714$\pm$0.12 $\vert$ 0.809$\pm$0.13}} & \multicolumn{1}{c|}{\textbf{0.676$\pm$0.2 $\vert$ 0.767$\pm$0.23}}  & \multicolumn{1}{c|}{\textbf{0.700$\pm$0.05 $\vert$ 0.808$\pm$0.07}}    \\ \hline                                   
	\end{tabular}}
\end{table}

\begin{figure}[t!]
	\centering
	\includegraphics[width=1.0\columnwidth]{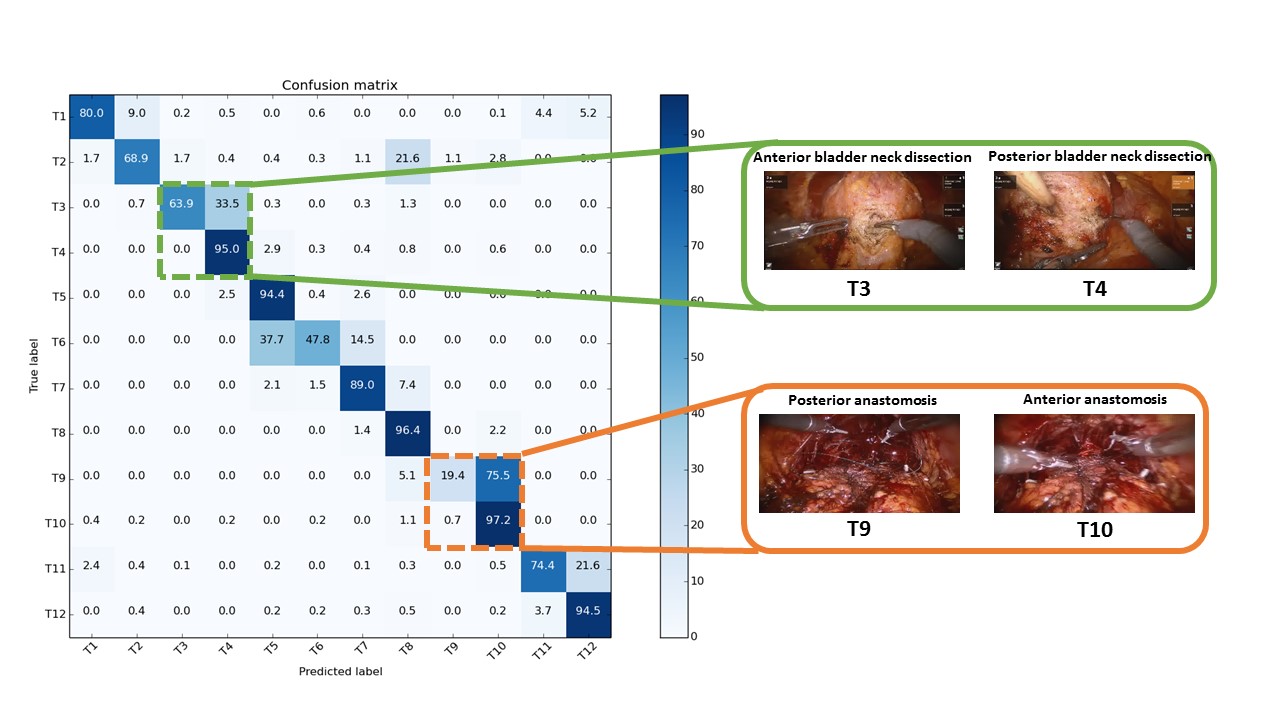}
	\caption{Confusion matrix of results using RP-Net with post-processing. Sample images of tasks between which there is a lot of \textit{`confusion'} are also shown.}
	\label{fig:confmat}
\end{figure}

Figure \ref{fig:confmat} shows the confusion matrix of RP-Net with post-processing. The model performed well for almost all the tasks individually except for task 9. However, we can see that most of the task 9 samples were classified as task 10. Tasks 9 and 10 are very related - they are two parts of one overall task (posterior and anterior anastomosis). Furthermore, the images from these two tasks were quite similar given they show anatomy during reconstruction after extensive dissection and energy application. Hence, one would expect that the model could be confused on these two tasks. This is also the case for tasks 3 and 4 - anterior and posterior bladder neck dissection, respectively. 

\begin{figure}[t!]
	\centering
	\includegraphics[width=1.0\columnwidth]{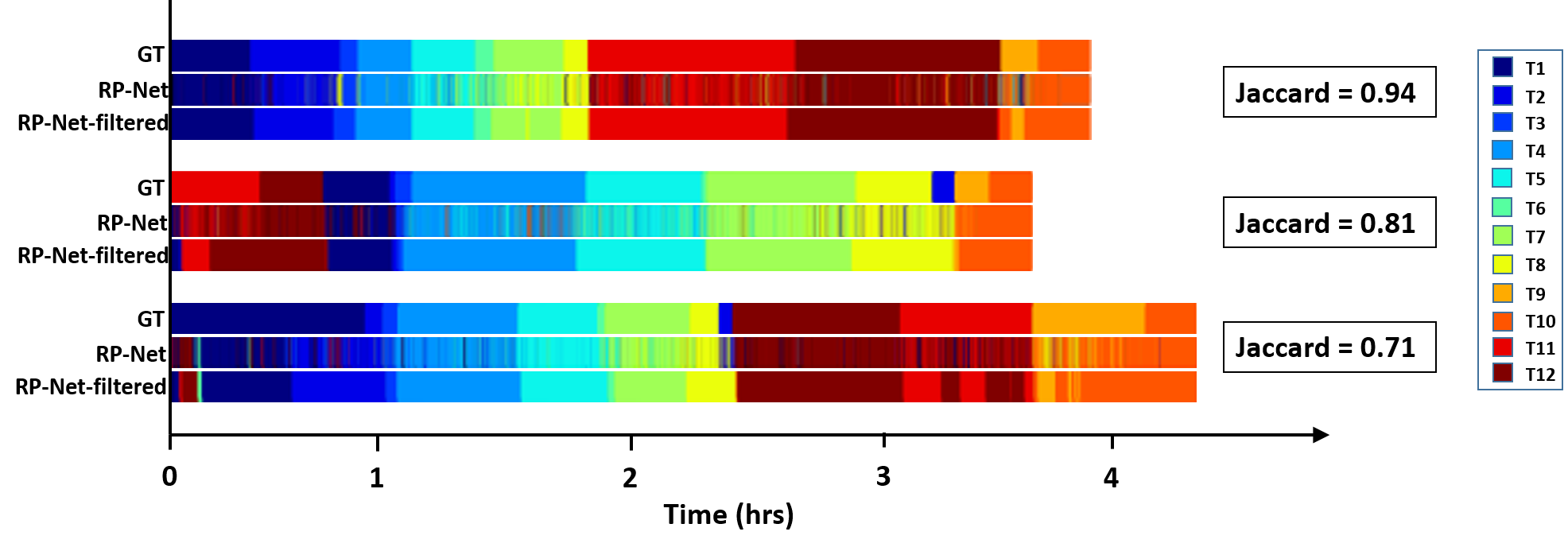}
	\caption{Sample segmentation outputs for the best, median and lowest jaccard index achieved (from top to bottom, respectively). Within each plot, the top bar denotes the ground truth, the middle one shows the output of RP-Net, while the lowest one shows the output after applying the median filter. Please see Table \ref{dataset} for task names.}
	\label{fig:seg_bars}
\end{figure}

Figure \ref{fig:seg_bars} shows several visualizations of the segmentation results as color-coded bars. Undesired spikes in the predicted surgical phase were present when using the output of RP-Net directly. This can be explained by the fact that the model has no temporal information and classifies only using a single image which can lead to mis-classifications since different tasks can look similar at certain points in time. However, using the proposed median filter for post-processing significantly remove such noise and produces a more consistent output (compare middle to bottom bars for all three sample segmentation outputs in Figure \ref{fig:seg_bars}).

Despite not having temporal motion information, single image-based models recognize surgical tasks quite well. One reason for this result could be due to the significantly large dataset available for single-image based models. Given the presented RNN and C3D models use a window from the overall task as input, the amount of training data available for such models reduces by a factor of the length of window segment. Additionally, the RNN models might not have performed as well as similar work because in this work we recognized gross tasks directly whereas prior work focused on sub-task gestures and/or maneuvers \cite{dipietro2016recognizing}. Finally, C3D models remain difficult to train. Improved training of these models could lead to better results, which aligns with the intuition that temporal windows of image frames could provide relevant information for activity recognition.

\section{Conclusion}

In this paper, we proposed a deep learning model called RP-Net to recognize the steps of robot-assisted radical prostatectomy (RARP). We used a clinically-relevant dataset of 100 RARPs from one academic center which enables translation of our models to directly impact real-world surgeon training and medical research. In general, we showed that image-based models outperformed models using only surgeon motion and event data. In future work, we plan to develop novel models that optimally combine motion and image features while using larger dataset and to explore how our models developed for RARP extend to other robot-assisted surgical procedures.

\bibliographystyle{splncs}
\bibliography{MICCAI_2018}   
\end{document}